\documentclass[letterpaper]{article} 
\usepackage{aaai23}  
\usepackage{times}  
\usepackage{helvet}  
\usepackage{courier}  
\usepackage[hyphens]{url}  
\usepackage{graphicx} 
\usepackage{natbib}  
\usepackage{caption} 
\usepackage{algorithm}
\usepackage{algorithmic}

\usepackage{extarrows}
\usepackage{amsthm,amsmath,amssymb}
\usepackage{color,soul}
\usepackage{booktabs}
\usepackage{multirow}
\usepackage{array}
\usepackage{makecell}
\usepackage{ragged2e}
\usepackage{subfigure} 
\nocopyright

\usepackage{newfloat}
\usepackage{listings}
\DeclareCaptionStyle{ruled}{labelfont=normalfont,labelsep=colon,strut=off} 
\floatstyle{ruled}
\newfloat{listing}{tb}{lst}{}
\floatname{listing}{Listing}
\title{T2-GNN: Graph Neural Networks for Graphs with Incomplete Features and Structure via Teacher-Student Distillation}
\author{
    Cuiying Huo\textsuperscript{\rm 1},
    Di Jin\textsuperscript{\rm 1,2},
    Yawen Li\textsuperscript{\rm 3},
    Dongxiao He\textsuperscript{\rm 1,}\thanks{Corresponding author},
    Yu-Bin Yang\textsuperscript{\rm 2},
    Lingfei Wu\textsuperscript{\rm 4}
}
\affiliations{
    \textsuperscript{\rm 1}College of Intelligence and Computing, Tianjin University, Tianjin, P.R. China\\
    \textsuperscript{\rm 2}State Key Laboratory for Novel Software Technology, Nanjing University, Nanjing, P.R. China\\
    \textsuperscript{\rm 3}School of Economics and Management, Beijing University of Posts and Telecommunications, Beijing, P.R. China\\
    \textsuperscript{\rm 4}Pinterest, New York, USA\\
    
    \{huocuiying,  jindi, hedongxiao\}@tju.edu.cn, warmly0716@bupt.edu.cn, yangyubin@nju.edu.cn, lwu@email.wm.edu
}

\usepackage{bibentry}

\begin{document}
\maketitle

\begin{abstract}
Graph Neural Networks (GNNs) have been a prevailing technique for tackling various analysis tasks on graph data. A key premise for the remarkable performance of GNNs relies on complete and trustworthy initial graph descriptions (i.e., node features and graph structure), which is often not satisfied since real-world graphs are often incomplete due to various unavoidable factors. In particular, GNNs face greater challenges when both node features and graph structure are incomplete at the same time. The existing methods either focus on feature completion or structure completion. They usually rely on the matching relationship between features and structure, or employ joint learning of node representation and feature (or structure) completion in the hope of achieving mutual benefit. However, recent studies confirm that the mutual interference between features and structure leads to the degradation of GNN performance. When both features and structure are incomplete, the mismatch between features and structure caused by the missing randomness exacerbates the interference between the two, which may trigger incorrect completions that negatively affect node representation. To this end, in this paper we propose a general GNN framework based on teacher-student distillation to improve the performance of GNNs on incomplete graphs, namely T2-GNN. To avoid the interference between features and structure, we separately design feature-level and structure-level teacher models to provide targeted guidance for student model (base GNNs, such as GCN) through distillation. Then we design two personalized methods to obtain well-trained feature and structure teachers. To ensure that the knowledge of the teacher model is comprehensively and effectively distilled to the student model, we further propose a dual distillation mode to enable the student to acquire as much expert knowledge as possible. Extensive experiments on eight benchmark datasets demonstrate the effectiveness and robustness of the new framework on graphs with incomplete features and structure.
\end{abstract}

\section{Introduction}

Graph Neural Networks (GNNs), as a powerful technique for modeling and analysis of graphs, have gained widespread attention in recent years as the graph-structured data become ubiquitous~\cite{GNNBook2022}. GNNs achieve remarkable results across a wide range of graph analysis tasks such as node classification~\cite{DBLP:conf/kdd/0017ZB0SP20, DBLP:conf/icdm/YuJLHWT021}, link prediction~\cite{DBLP:conf/icml/YouYL19, DBLP:conf/kdd/ChenYS0GM20}, and recommendation~\citep{DBLP:conf/cikm/XuLHLX019, DBLP:conf/sdm/LiuWGAY20}, and are continuously being applied in many fields such as computer vision and natural language processing~\cite{gnnsurvey1,DBLP:journals/aiopen/ZhouCHZYLWLS20,jin-survey}. The essence of most existing GNNs is the process of message propagation and aggregation guided by graph structure (neighbors), such as GCN~\cite{DBLP:conf/iclr/KipfW17}, GraphSAGE~\cite{DBLP:conf/nips/HamiltonYL17}, GAT~\cite{DBLP:conf/iclr/VelickovicCCRLB18} and APPNP~\cite{DBLP:conf/iclr/KlicperaBG19}. Therefore, a key prerequisite for the success of GNNs is complete and trustworthy initial graph descriptions (i.e., node features and graph structure).

However, in real-world scenarios, there are many graph data with incomplete descriptions for various reasons. For example, some features or edges are missing due to technical or human errors during data collection. The large scale and potential resource constraints of complex real-world systems make collecting complete data prohibitively expensive or impossible. And we cannot obtain some sensitive feature information due to confidential data or privacy security (e.g., nationality, age). Furthermore, the addition of a new node leads to the absence of links between the node and other nodes in the graph, and even becomes an isolated node. All these cases lead to incomplete features or structure in the initial graph, which brings great challenges to GNNs. 

\begin{figure}[t]  
	\centering       
	\includegraphics[width=0.99\linewidth]{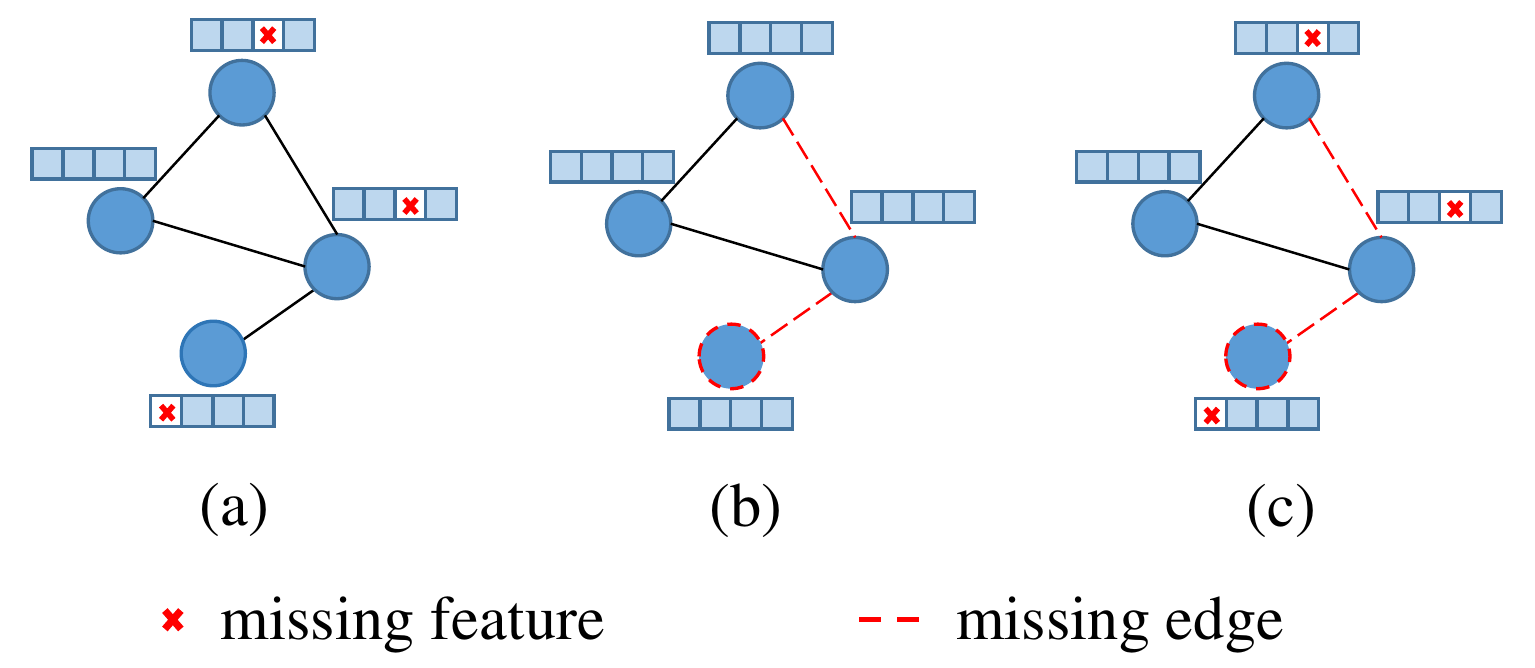}
	\caption{Three types of incomplete data. (a) Feature-incomplete graph.
		(b) Structure-incomplete graph. (c) Incomplete graph.}
	\label{fig:im}    
\end{figure}

Based on the types of missing descriptions in the graphs, descriptions missing graphs can be classified into three categories: 1) feature-incomplete graph where features of some nodes are partially (or entirely) missing, as shown in Fig.~\ref{fig:im}(a); 2) structure-incomplete graph where edges of some nodes are partially (or entirely) missing (Fig.~\ref{fig:im}(b)); 3) incomplete graph where features and (or) edges of some nodes are partially (or entirely) missing (Fig.~\ref{fig:im}(c)). For GNN-based methods for feature-incomplete graphs, some specific methods have been proposed. SAT~\cite{DBLP:journals/pami/ChenCYZZT22} proposes a structure-feature matching GNN model to complete the missing features by assuming that the graph structure and node features come from the same latent space. GCNMF~\cite{DBLP:journals/fgcs/TaguchiLM21} integrates the completion of missing features and graph learning within the same neural network architecture by representing the missing data with a Gaussian mixture model. HGNN-AC~\cite{hgnn-ac} uses graph structure information as a priori guidance and proposes an end-to-end attribute completion framework for heterogeneous graphs. There are also many attempts to design GNN algorithms to model the structure-incomplete graphs by jointly learning optimized graph structure and node representations~\cite{DBLP:conf/nips/0022WZ20,DBLP:journals/corr/abs-2111-04840,DBLP:journals/corr/abs-2103-03036}. But to our best knowledge, no method has been proposed for incomplete graphs, which is more challenging and relevant to many real-world applications. 

In addition, these existing feature (or structure) completion methods usually rely on the matching relationship between features and structure to guide feature (or structure) completion, or employ the joint learning of node representation and information completion in the hope of achieving reciprocity between the two. However, recent studies have confirmed that the mutual interference between features and structure in graphs is an important reason for the performance degradation of GNNs~\cite{DBLP:conf/www/0002ZPNG0CH22,DBLP:conf/kdd/0017ZB0SP20}. Due to the randomness of missing, features and structure usually cannot be completely matched in the graph, which makes features and structure interfere with each other in the message passing of GNN. In incomplete graphs, both features and structure are incomplete, which may make the mutual interference between the two more serious. Therefore, the methods based on the above mechanisms cannot be effectively applied to incomplete graphs, and even aggravate the propagation of error information, resulting in completed information having a negative impact on node representation.

To this end, in this paper, we introduce knowledge distillation to address the above problems, and propose a new general GNN framework (T2-GNN) for incomplete graphs. The new T2-GNN framework addresses three key limitations: (i) How to improve the performance of the base GNN (e.g., GCN) model in the case of incomplete features and graph structure, and efficiently avoid potential negative interference between features and structure? (ii) Given incomplete node features and graph structure, how can we effectively model them in favor of learning better node representations? (iii) How to distill the knowledge of the teacher model into the student model to achieve effective knowledge transfer?

For the first limitation, we design teacher models based on node features and graph structure respectively to avoid potential interference between the two. Then, based on knowledge distillation, the well-trained teacher models are used to provide feature-level and structure-level targeted guidance for the message passing process of the student model (base GNN) to improve the performance of the student model on incomplete graphs. For the second limitation, in the feature teacher, we design parameterized imputation for feature completion, and adopt joint learning to capture effective information at the feature level. In the structure teacher, we construct a structure-enhanced graph in a personalized way to capture effective information at the structure level. Finally, for the third limitation, we propose a dual distillation mode to achieve comprehensive guidance from the teacher model to the student model.

Note that, compared with traditional knowledge distillation, our aim is not to distill knowledge from a cumbersome teacher model into a lightweight student model so that the student can hold a similar performance as the teacher’s. Instead, our aim is for the issues existing in the student model, the teacher models can provide targeted and effective guidance, i.e., suit the medicine to the illness. In short, we summarize the main contributions as follows:
\begin{itemize}
\item We investigate a novel problem that existing GNNs fail to perform well when both features and structure of the graph are incomplete. And we find that the mutual interference between features and structure makes it more challenging to effectively model incomplete graphs.
\item We propose a general GNN framework (T2-GNN) for incomplete graphs through teacher-student distillation,  which introduces feature-level and structure-level teacher models to provide targeted guidance for student (base GNN), and designs a dual distillation mode to enhance the effective transfer of knowledge. It can improve the performance of the base GNN while avoiding the mutual interference of features and structure, and is easy to be combined with an arbitrary base GNN.
\item By combining with four classical base GNNs, our framework consistently achieves performance improvements across eight benchmark datasets. Experiments comparing with different kinds of baselines further verify the robustness of the new framework.
\end{itemize}

\section{Preliminaries}
\subsection{Notations}
Given an undirected graph $G=(V,X)$, where $V=\{v_1,v_2,...,v_n\}$ is the set of $n$ nodes, $X  \in \mathbb{R}^{n\times d}$ denotes the feature matrix and $d$ is the feature dimension. We let $A = [a_{ij}] \in \{0,1\}^{n\times n}$ represents the adjacency matrix, where $a_{ij}=1$ denotes there exists an edge between nodes $v_i$ and $v_j$, or 0 otherwise. An incomplete graph means that some elements $x_{ij}$ in the feature matrix $X$ are missing, and $a_{ij}$ in the adjacency matrix $A$ should be set to 1 and set to 0 by mistake. In this paper, we focus on the semi-supervised node classification task, which means that only a subset of nodes ($V_L, |V_L|\ll n$) have ground truth labels $Y$. Our task is to predict the labels of $V \backslash V_L$.

\subsection{Graph Neural Networks}
Graph Neural Networks (GNNs) have shown great power in tackling various analysis tasks on graph-structured data. Most of the existing GNNs follow the message-passing mechanism, that is, propagating and aggregating information guided by the graph structure (neighbors), such as GCN, GAT, etc. Formally, in the process of GNN learning node representations, the representation of node $u$ at the $l^{th}$ layer can be formalized as follows:
\begin{equation}\label{eq:gnn-prop}
    p_{v}^{(l)}=\text{Pro}{{\text{p}}^{(l)}}\left( x_{v}^{(l-1)} \right),\quad v\in \mathcal{N}\left( u \right),
\end{equation}
\begin{equation}\label{eq:gnn-agg}
    x_{u}^{(l)}=\text{Ag}{{\text{g}}^{(l)}}\left( x_{u}^{(l-1)},\{p_{v}^{(l)}:v\in \mathcal{N}\left( u \right)\} \right),
\end{equation}
where $\mathcal{N}\left( u \right)$ is the set of the neighbors of node $u$, $\text{Pro}{{\text{p}}^{(k)}}\left( \cdot  \right)$ is the node representation transformation in the propagation process and $\text{Ag}{{\text{g}}^{(k)}}\left( \cdot  \right)$ aggregates information of its neighborhood and itself to obtain the final node representation. Both $\text{Pro}{{\text{p}}^{(k)}}\left( \cdot  \right)$ and $\text{Ag}{{\text{g}}^{(k)}}\left( \cdot  \right)$ are functions implemented by neural networks in the $l^{th}$ convolutional layer of GNN. Given $u$’s input features $x^0$ and its neighborhoods $\mathcal{N}\left( u \right)$, we can use Eq.(\ref{eq:gnn-prop}) and Eq.(\ref{eq:gnn-agg}) to obtain its final representation.

However, the initial features and graph structure of real-world graphs are often incomplete,
which will make GNNs using the above propagation and aggregation functions learn the suboptimal node representations. In this paper, we focus on improving the performance of GNNs when both features and graph structure are incomplete. 

\subsection{Knowledge Distillation}
Knowledge distillation aims to distill knowledge from a cumbersome teacher model into a lightweight student model so that the student can hold a similar performance as the teacher's. Vanilla knowledge distillation was first proposed by Hinton~\cite{DBLP:journals/corr/HintonVD15}, which provides softened softmax labels through the teacher model to guide the training of the student model. Given the output logits $Z_T$ and $Z_S$ of the teacher model and the student model, the distillation loss is as follows:
\begin{equation}\label{eq4l}
    \mathcal{L}_{KD} =\mathcal{H}(\sigma_S{(Z_S;\rho)},\sigma_T{(Z_T;\rho)}),
\end{equation}
where $\mathcal{H}$ is negative cross-entropy loss or KL-divergence, $\sigma$ denotes the softmax with temperature $\rho$. The larger the $\rho$, the smoother the output probability, the information carried by the negative labels will be relatively amplified, and the model training will pay more attention to the negative labels.

In addition, some distillation models based on intermediate representations are constantly being proposed. They achieve knowledge distillation by aligning intermediate representations (such as attention map~\cite{DBLP:conf/iclr/ZagoruykoK17}, intermediate features~\cite{DBLP:journals/corr/HuangW17a}) between the teacher model and the student model. Common alignment functions such as $L_2$ loss:
\begin{equation}
\mathcal{L}_{2} =||R_S-R_T||_2^2,
\end{equation}
where $R_S$ and $R_T$ are the intermediate representations of the student and teacher models, respectively.

To the best of our knowledge, we employ knowledge distillation for the first time to improve the robustness of GNNs on incomplete graphs, effectively avoiding inaccurate feature completion or structure enhancement that is exacerbated by the potential interference between features and graph structure.
\begin{figure}[t]  
	\centering       
	\includegraphics[width=0.99\linewidth]{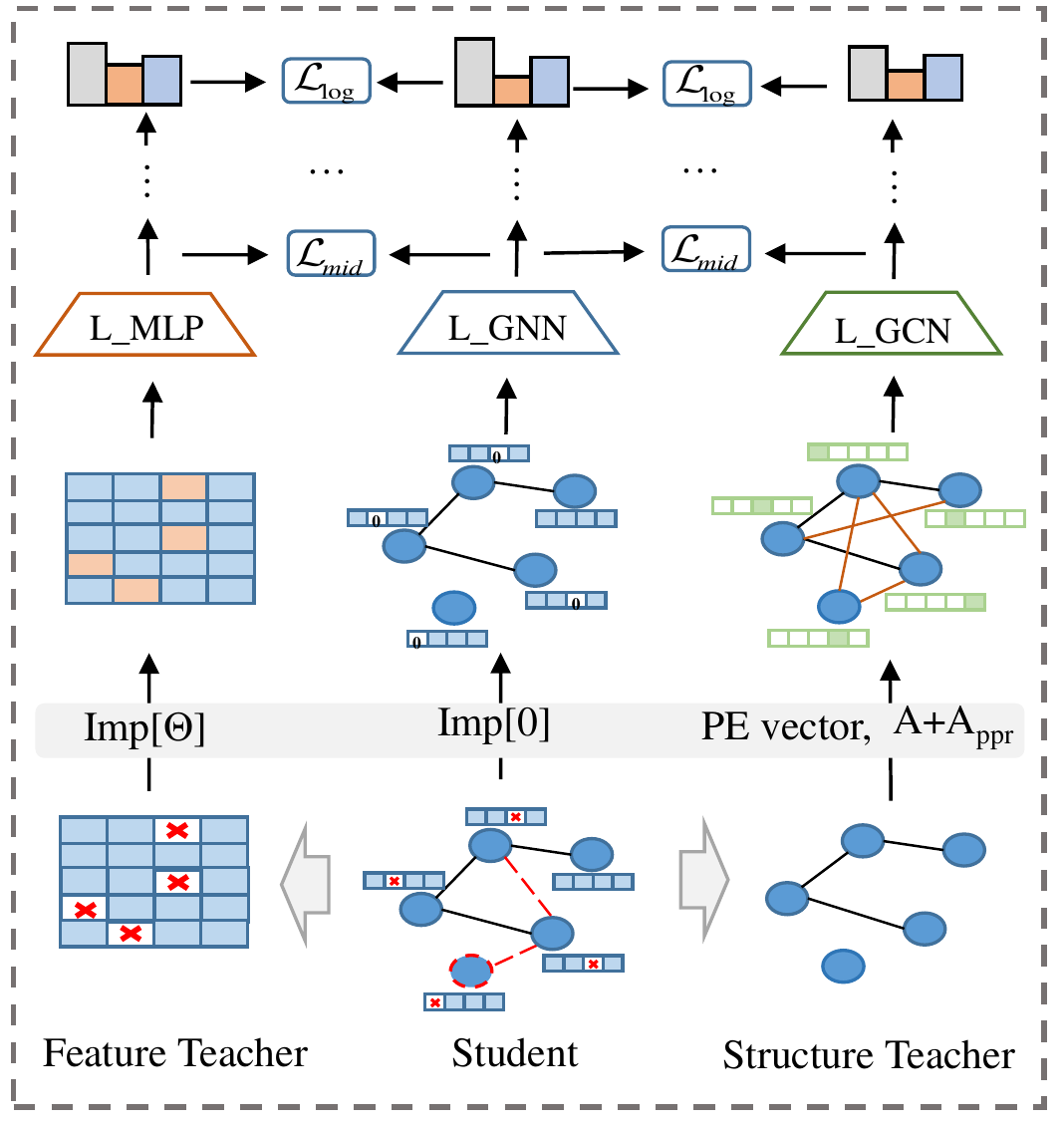} 
	\caption{Overview of the T2-GNN framework.}\label{fig:datu}  
\end{figure}

\section{T2-GNN Framework}
\subsection{Overview}
To address the problem that GNNs fail to perform well on incomplete graphs, we propose a novel and general GNN framework based on teacher-student distillation that can effectively model both incomplete features and graph structure simultaneously and improve the robustness of GNNs, namely T2-GNN. The new framework aims to design personalized feature teacher and structure teacher models on incomplete graphs, and provide targeted and effective guidance for student model (such as GCN) and avoid interference between features and structure through knowledge distillation. It has three components: feature teacher with parameterized completion, structure teacher with personalized enhancement, as well as a dual distillation module (Fig.~\ref{fig:datu}). The feature teacher with parameterized completion is designed to capture valuable information in incomplete features while avoiding the potential interference of graph structure on its learning process. It assigns parameterized imputation to missing features, uses multi-layer perceptrons to model node features, and conducts joint training for feature completion and node representation. The structure teacher with personalized enhancement is designed to effectively model the incomplete graph structure while avoiding the potential interference of node features during model learning. It uses personalized PageRank to enhance the incomplete initial graph structure, and replaces the incomplete initial node features with positional encoding vector to further enhance the personalized structure information of nodes. Then it feeds the new structure augmented graph into GCN for model training. The dual distillation module is designed to maximize the distillation of targeted teachers' expert knowledge into the student model to effectively improve its performance on incomplete graphs. It adopts both contrast-based intermediate representation alignment and logit-based soft label-guided dual distillation mode to achieve effective and comprehensive transfer of teacher's knowledge to student. Finally, the student model is jointly optimized by the distillation losses and the original classification loss.

\subsection{Feature Teacher with Parameterized Completion}
Previous GNN methods for modeling incomplete node features usually rely on pre-defined assumptions for feature completion. For example, SAT~\cite{DBLP:journals/pami/ChenCYZZT22} assumes that the graph structure is complete and use the matching relationship between structure and feature for feature completion, GCNMF~\cite{DBLP:journals/fgcs/TaguchiLM21} assumes that all features follow a same data distribution and use distribution parameters for feature imputation. However, such strong assumptions are not universal because real-world graph data are usually complex and diverse, and may also be accompanied by incomplete graph structure. In addition, the potential mutual interference between features and graph structure in the learning process of GNN has also been confirmed in recent studies~\cite{DBLP:conf/www/0002ZPNG0CH22,DBLP:conf/kdd/0017ZB0SP20}. 

To this end, we propose to use a parameterized completion mechanism to complete the missing features through the learning ability of the model itself, so as to achieve the reciprocity of feature completion and model learning. Specifically, given an incomplete feature matrix $X  \in \mathbb{R}^{n\times d}$, we use learnable neural network parameters to impute missing features. The feature matrix for parameterized imputation can be formalized as:
\begin{equation}
\overline{X}_{ij}=
\begin{cases}
\Theta_{ij}& \text{  if $X_{ij}$ is missing;} \\
X_{ij}& \text{  otherwise,}
\end{cases}
\end{equation}
where $\Theta \in \mathbb{R}^{n\times d}$ is a randomly initialized neural network parameter matrix.

In order to fully capture the valuable information in the features while avoiding the potential interference caused by the graph structure, we only utilize node features to construct feature teacher model. We employ a multi-layer perceptron (MLP) to model the features. Thus we can obtain the output logits $Z_{fea}$ and intermediate representations $R_{fea}$ of the feature teacher for distillation:
\begin{equation}
    Z_{fea}, R_{fea}=f_{mlp}(\overline{X}),
\end{equation}
where $f_{\text{mlp}}(\cdot)$ is the MLP with the ReLU activation function. Its training objective function can be written as:
\begin{equation}\label{eq7}
   \underset{\Theta_{fea}}{\text{min}} = \sum\nolimits_{{v_i} \in {{\cal T}_{\cal V}}} {\mathcal{J}_c\left( {{\sigma({z}_{fea,i}}),{y_i}} \right)},  
\end{equation}
where $\Theta_{fea}$ is all parameters of the feature teacher including imputation parameter $\Theta$, $\sigma(\cdot)$ the softmax function, ${{\mathcal{T}}_{{V}}}$ the nodes in training set, $\mathcal{J}\left( \cdot  \right)$ the cross entropy loss, and ${{y}_{i}}$ the true one-hot label of node ${{v}_{i}}$.

\subsection{Structure Teacher with Personalized Enhancement}
Neighbor nodes are crucial for the target node to learn representations, but the real-world graph structure is usually incomplete, so that some neighbors of the target node are missing, and even becomes an isolated node. Therefore, we first adopt personalized PageRank, which is very effective in expanding important neighbors in the graph~\cite{DBLP:conf/iclr/KlicperaBG19}, to augment the incomplete graph structure. 

Personalized PageRank (PPR)~\cite{DBLP:conf/www/Haveliwala02} takes the graph structure as input, adopts a restarted random walk strategy to sample neighbors for the target node $v_i$, and calculates the ranking score $p_{ij}$ between $v_i$ and the sampled node $v_j$. A larger $p_{ij}$ indicates that the two nodes are more related. Similar to ~\cite{DBLP:conf/kdd/BojchevskiKPKBR20}, we use a push iteration method to compute ranking scores, which can be approximated effectively even for very large networks. Formally, the PPR matrix $P$ is calculated as:
\begin{equation}\label{eq17}
    P =  \left( {1 - \alpha } \right)\hat{A}P + \alpha I,
\end{equation}
where $\alpha$ is the reset probability, $\hat{A}=A+I$. Then, since there are inevitably some unimportant scores in $P$, we only keep the top $k$ important scores for each node in the $P$ and set the other elements to zero to get the PPR-based structure matrix $A_{ppr}$.

Considering that the initial graph structure still carries rich and useful information, we combine the initial adjacency matrix $A$ with the PPR-based matrix $A_{ppr}$ to construct the enhanced graph structure: $\overline{A}= A + A_{ppr}$.

Different from the feature teacher only uses node features for model learning, we adopt positional encoding (PE) vectors to replace the initial incomplete features to construct a complete personalized enhancement graph for structure teacher. PE can effectively model position information in sequence data~\cite{DBLP:conf/nips/VaswaniSPUJGKP17}. Here we intend to design the PE vectors to further extract the personalized (location) information of nodes in the graph structure. We use one-hot encoding to compute PE vectors, and in order to improve the scalability of PE vectors in large-scale graph data, we use a linear transformation to transform the vectors into the same dimension as the initial node feature. Formally, the PE vector of node $v$ is:
\begin{equation}\label{eq:attr-trans-node}
    {PE(v)}={{W}}\cdot {{x}_{v}} + b,
\end{equation}
where $W$ and $b$ are learnable parameter matrix and bias vector, ${x}_{v}$ the one-hot vetor of node $v$. 

Then we feed the complete enhancement graph into multi-layer GCN to obtain the output logit and intermediate representation of the structure teacher for distillation.
\begin{equation}
    Z_{str}, R_{str}=f_{gcn}(\overline{A},PE),
\end{equation}
where $f_{\text{gcn}}(\cdot)$ is the GCN with multiple convolutional layers. Its training objective function is the same as that of the feature teacher (Eq.(\ref{eq7})).

\subsection{Dual Distillation and Model Training}
After obtaining the output logits and intermediate representations of the structure and feature teachers, we design a dual distillation mode to provide targeted and effective guidance for the student model to the greatest extent. The output logits are demonstrated to contain rich information~\cite{DBLP:journals/corr/HintonVD15}, so we first use a logit-based distillation to achieve the knowledge transfer from the teacher model to the student model. Without loss of generality, we adopt KL-divergence as the distillation loss: 
\begin{equation}\label{eq4lo}
    \mathcal{L}_{log} =\mathcal{H}_{KL}\left(softmax(\frac{Z_s}{\rho}) || softmax(\frac{Z_t}{\rho})\right),
\end{equation}
where $Z_s$ is the output logits of the student model, $Z_t$ the output logits of the feature or structure teacher model, and $\rho$ the temperature parameter that controls the smoothness of the output.

The logit-based distillation can be considered to provide more label supervision information for student model by generating soft labels. Considering the negative effects of missing features and edges on the training process of student model, we further introduce intermediate representation distillation to guide the training of student model, which can also be considered as an indirect completion to missing features or edges in the training process of student model. We adopt a contrast-based intermediate representation alignment mechanism to achieve targeted guidance from teachers to student training by maximizing the consistency of intermediate embedding from the teacher model and the student model for the same node $i$ (i.e., positive samples ($r_i^s,r_i^t$)), and minimizing the consistency of the intermediate representation of $i$ in the student model and the intermediate representations of other nodes in the teacher model (i.e., negative samples ($r_i^s,r_j^t$)). The alignment loss is formalized as: 
\begin{equation}\label{eq:contrastiveInOneViewg}
    \mathcal{L}_{mid}(R_s,R_t)=\frac{1}{|{V}|}\sum\limits_{{{v}_{i}}\in {V}}{\log \frac{{{{e}^{\left\langle {r^s_i},{r^t_i} \right\rangle }}}}{{{{e}^{\left\langle {r^s_i},{{r^t_i}} \right\rangle }}+{\sum\nolimits_{j\neq{i}}{{e}^{\left\langle {r^s_i},{r^t_j} \right\rangle }}}}}},
\end{equation}
where $\left\langle {\cdot , \cdot} \right\rangle =\cos (\cdot , \cdot)/\rho' $, and $\rho'$ is a scaler temperature parameter. Then, we combine the above two losses as the final distillation loss between the student model and an arbitrary teacher model: $\mathcal{L}_{dual} = \mathcal{L}_{log} + \mathcal{L}_{mid}$.

Considering that the missing rates of features and edges in real-world graphs may be different, we adopt the parameter $\lambda$ to balance the two distillation losses. The overall training loss of the student model is as follows:
\begin{equation}\label{eq:contrastiveInOneViewd}
\mathcal{L}_{final}^{stu} = \mathcal{J}_c^{stu} + \lambda \mathcal{L}_{dual}^{fea} + (1-\lambda) \mathcal{L}_{dual}^{str}
\end{equation}
where  $\mathcal{J}_c^{stu}$ is standard cross-entropy loss, $\lambda$ the balance hyperparameter. 
\begin{table}[t]
	\centering
	\footnotesize
	\resizebox{0.99\linewidth}{!}{
		\begin{tabular}{c|c|c|c|c}
			\specialrule{1pt}{0pt}{0pt}
			\textbf{Datasets}  & Nodes & Edges  & Features &Classes  \\ \specialrule{0.5pt}{-0.8pt}{1pt}
			Texas        & 183       & 295     & 1703       & 5    \\
			Cornell      & 183       & 295     &1703        & 5      \\
			Wisconsin       &251        &499      &1703        &5       \\
			Chameleon      & 2,277         & 31,421      &2,325        &5       \\
			Cora     & 2,708         & 5,278     &1,433        &7       \\
			Citeseer    & 3,327         & 4,676 &  3,703      &6       \\
			Squirrel  & 5,201         & 198,493  & 2,089       & 5      \\
			Pubmed  & 19,717        & 44,327       &500        &3   
			\\ \specialrule{1pt}{0pt}{0pt}
		\end{tabular}
	}
	\caption{Statistics of datasets.}\label{tab:datasets}
\end{table}
\section{Experiments}
\subsection{Experimental setup}
\subsubsection{Datasets.}
We use eight real-world datasets of different scales for model evaluation (Table~\ref{tab:datasets}). They are two Wikipedia networks Chameleon and Squirrel~\cite{DBLP:journals/compnet/RozemberczkiAS21}, three webpage networks Cornell, Texas and Wisconsin\footnote{http://www.cs.cmu.edu/afs/cs.cmu.edu/project/theo-11/www/wwkb}, and three citation networks Cora, CiteSeer and PubMed~\cite{DBLP:journals/aim/SenNBGGE08,2012Query}. 

\subsubsection{Baselines.}
We first plug four GNNs as student models into T2-GNN, including GCN~\cite{DBLP:conf/iclr/KipfW17}, GAT~\cite{DBLP:conf/iclr/VelickovicCCRLB18}, GraphSAGE~\cite{DBLP:conf/nips/HamiltonYL17}, and APPNP~\cite{DBLP:conf/iclr/KlicperaBG19}, to verify whether T2-GNN can boost the performance of node classification. 

Then we compare our T2-GNN with two GNNs for features completion (SAT~\cite{DBLP:journals/pami/ChenCYZZT22} and GCNMF~\cite{DBLP:journals/fgcs/TaguchiLM21}) and two GNNs for structure enhancement (IDGL~\cite{DBLP:conf/nips/0022WZ20} and PTDNet~\cite{DBLP:conf/wsdm/LuoCYZNCZ21}). For a fair comparison, all methods use GCN as the backbone.

Finally, we use different training ways to construct two new variant models to further verify the effectiveness of T2-GNN: (1) combining feature completion and structure enhancement into a single-teacher model to guide student' training, named singleT; (2) conducting a joint end-to-end training of feature teacher, structure teacher and student model, instead of training the teacher models alone, named T2-GNN (online).
\subsubsection{Implementation Details.}
For all datasets, we randomly split nodes of each class into 60\%, 20\%, and 20\% for training, validation and testing, and generate 10 random splits for all methods. Since the missing degree of features and edges in real-world datasets is agnostic, we randomly mask 30\% of features and edges in each dataset for model evaluation. All parameters in the baselines adopt the original parameters in their paper. In our new framework T2-GNN, for feature teacher, we use a two-layer MLP with 512 units in the hidden layer. For structure teacher, we adopt GCN with two graph convolutional layers, and search the top $k$ for PPR-based structure enhancement strategy in \{5, 10, $\cdot\cdot\cdot$, 25\}. We set learning rate to 0.001 and weight decay to 0.0005 for both teacher models. For student model, we use its original parameter settings in the paper. We search the temperature parameter $\rho$ in the distillation loss in \{1, 2, $\cdot\cdot\cdot$, 5\} and the balance parameter $\lambda$ in \{0.1, 0.2, $\cdot\cdot\cdot$, 0.9\}.  
\vspace{-49pt}

\begin{center}
	\begin{table*}[t!]
		\centering
        \small
		\resizebox{0.78\linewidth}{!}{
			\begin{tabular}{c|ccccccccc} 
				\specialrule{1pt}{0pt}{0pt}
				\textbf{Datasets} & Texa. & Corn. & Wisc.  & Cham.   & Cora   & Cite. & Squi.  &Pubm. & avg \\ \specialrule{0.5pt}{-0.8pt}{1pt}
				GCN   &46.48  &48.90  &50.39   &58.37  &82.77   &68.84   &40.09    &83.51 &59.90\\
				T2-GCN    &65.40  &67.29  &73.39   &60.81  &83.84   &72.32   &44.33   &84.85 &69.02 \\\specialrule{0.5pt}{-0.8pt}{1pt}
				\textbf{+Impv.} &18.92  &18.39  &23.00   &2.44  &1.07   &3.58   &4.24   &1.34 &9.12  \\ \specialrule{0.5pt}{-0.3pt}{1pt} \specialrule{0.5pt}{-0.3pt}{1pt}
				GAT &50.27  &55.67  &53.72   &57.56  &83.54   &70.00   &36.58    &84.33 &61.45\\
				T2-GAT  &61.35  &64.86  &62.35   &58.39  &84.46   &72.42   &38.87   &85.07  &65.96\\\specialrule{0.5pt}{-0.8pt}{1pt}
				\textbf{+Impv.}  &11.08  &9.19  &8.63   &0.83  &0.92   &2.42   &2.33    &0.74  &4.51 \\ \specialrule{0.5pt}{-0.3pt}{1pt} \specialrule{0.5pt}{-0.3pt}{1pt}
				GraphSAGE  &70.81  &64.59  &69.41   &59.05  &82.93   &70.96  &41.86    &84.96  &68.07\\
				T2-SAGE   &78.10  &75.40  &82.54   &62.89  &84.62   &71.39   &45.78
				&85.56  &73.28\\\specialrule{0.5pt}{-0.8pt}{1pt}
				\textbf{+Impv.}  &7.29  &10.81  &13.13   &3.84  &1.69   &0.43   &3.92   &0.60  &5.21 \\ \specialrule{0.5pt}{-0.3pt}{1pt} \specialrule{0.5pt}{-0.3pt}{1pt}
				APPNP  &47.83  &55.94  &53.52   &44.18  &83.48   &70.61   &28.57    &78.77 &57.86 \\
				T2-APPNP    &57.83  &66.48  &66.27   &50.41  &84.08   &71.77   &36.93   &83.36  &64.63\\\specialrule{0.5pt}{-0.8pt}{1pt}
				\textbf{+Impv.}  &10.00  &10.54  &12.75   &6.23  &0.60   &1.16   &8.36   &4.59  &6.77 \\\specialrule{1pt}{0pt}{0pt}
			\end{tabular}
		}
		\caption{Comparisons on node classification (Percent). "+Impv" is the performance improvement of T2-GNN for base GNN.}\label{tab:clustering}
	\end{table*}
\end{center}
\begin{center}
	\begin{table*}[t]
		\centering
        \small
		\resizebox{0.78\linewidth}{!}{
			\begin{tabular}{c|ccccccccc}
				\specialrule{1pt}{0pt}{0pt}
				\textbf{Datasets} & Texa. & Corn. & Wisc.  & Cham.   & Cora   & Cite. & Squi.  &Pubm. & avg \\ \specialrule{0.5pt}{-0.8pt}{1pt}
				GCN   &46.48  &48.90  &50.39   &58.37  &82.77   &68.84   &40.09    &\ul{83.51} & 59.90\\\specialrule{0.5pt}{-0.3pt}{1pt} \specialrule{0.5pt}{-0.3pt}{1pt}
				SAT &\ul{63.89} &\textbf{76.68} &67.49  &\ul{56.57} &64.59 &45.87 &35.23  &- &58.61  \\
				GCNMF &57.28 &55.43 &53.02 &47.75  &\ul{83.08} &71.75 &30.56  &67.08 &58.24\\\specialrule{0.5pt}{-0.8pt}{1pt}
				IDGL &62.16 &54.05 &60.78 &48.68  &79.88 &65.80 &33.11 &82.60 &60.88  \\
				PTDNet &61.62 &64.87 &\ul{73.14} &48.42  &74.93 &\ul{72.29} &30.45 &82.49 &63.52 \\\specialrule{0.5pt}{-0.8pt}{1pt}
				singleT &53.81 &54.76 &57.10 &55.93  &80.19 &67.22 &40.32  &81.03 &61.29  \\
				T2-GCN (online) &60.53 &\ul{74.01} &69.42 &54.22  &80.11 &65.31 &\textbf{45.19}  &83.04 &\ul{66.47}  \\
				T2-GCN    &\textbf{65.40}  &67.29  &\textbf{73.39}   &\textbf{60.81}  &\textbf{83.84}   &\textbf{72.32}   &\ul{44.33} &\textbf{84.85} &\textbf{69.02} \\\specialrule{1pt}{0pt}{0pt}
			\end{tabular}
		}
		\caption{Comparisons on node classification (Percent). The best results are in bold and the second best is underlined. "-": OOM.}\label{tab:clusterin}
	\end{table*}
\end{center}
\begin{center}
	\begin{table}[t]
		\centering
		\small
		\resizebox{0.9\linewidth}{!}{
			\begin{tabular}{c|ccccc}
				\specialrule{1pt}{0pt}{0pt}
				\textbf{Missing rate} & 0\% & 20\% & 40\%  & 60\%   & 80\%    \\ \specialrule{0.5pt}{-0.8pt}{1pt}
				GCN   &62.33 &60.82 &57.85 &52.43 &45.17\\
				T2-GCN  &71.30 &69.93 &66.54 &60.44 &54.60 \\\specialrule{1pt}{0pt}{0pt}
			\end{tabular}
		}
		\caption{Comparisons of the average accuracy on eight datasets under different missing rates (Percent).}\label{tab:mask}
	\end{table}
\end{center}


\subsection{Node Classification}
We report here the accuracy on the node classification task (Table~\ref{tab:clustering} and Table~\ref{tab:clusterin}), and compare the average accuracy of node classification on eight datasets under different features and edges missing rates (Table~\ref{tab:mask}). We have the following observations.

In Table~\ref{tab:clustering}, T2-GNN consistently improves the performance of the four backbones on all datasets. The average improvement is 9.12\%, 4.51\%, 5.21\% and 6.77\% on GCN, GAT, GraphSAGE and APPNP, respectively. In particular, T2-GNN has achieved significant performance improvement on small datasets (Texas, Cornell and Wisconsin). This is because it is well known that small datasets are more prone to over-smoothing problems than large datasets under the same network depth. However, the missing of features or edges may be accompanied by the missing of important and individualized features or structure of the node, thus aggravating the over-smoothing problem. T2-GNN individually models incomplete features and structure, and enhances the personalized nature of nodes in features and structure, respectively, which alleviates above problem to some extent. 

T2-GNN outperforms the baseline methods in most cases (Table~\ref{tab:clusterin}), which validates the necessity of GNN modeling incomplete features and graph structure simultaneously. Furthermore, GCN outperforms feature completion methods (SAT and GCNMF) and structure-enhanced methods (IDGL and PTDNet) in many cases. This is because these methods generally believe that the potential relationship between features and structure is beneficial to solve the problem of incomplete features (or structure). However, when both features and structure are incomplete, this mechanism may have a negative influence on model performance due to potential interference between features and structure. T2-GNN effectively avoids this negative influence by completing features and structure separately.

T2-GNN outperforms singleT and T2-GNN (online) in most cases, and even GCN can perform better than these two variant models on 4 out of 8 datasets. Although T2-GNN (online) also models incomplete features and structure separately, the mechanism of joint training still cannot effectively avoid the potential (indirect) interference between the two during training. T2-GNN trains each model in an individually optimized manner so that it is free from this interferences.

In addition, as shown in Table~\ref{tab:mask}, T2-GCN has a stable and significant performance improvement against GCN, with 8.97\%, 9.11\%, 8.69\%, 8.01\% and 9.43\% improvement (on average on all the datasets) under different masking ratios, which further validates the effectiveness and robustness of T2-GNN.

\begin{figure}[t]  
	\centering    
	\subfigure[Wisconsin] 
	{
	    \centering
		\includegraphics[width=0.45\columnwidth]{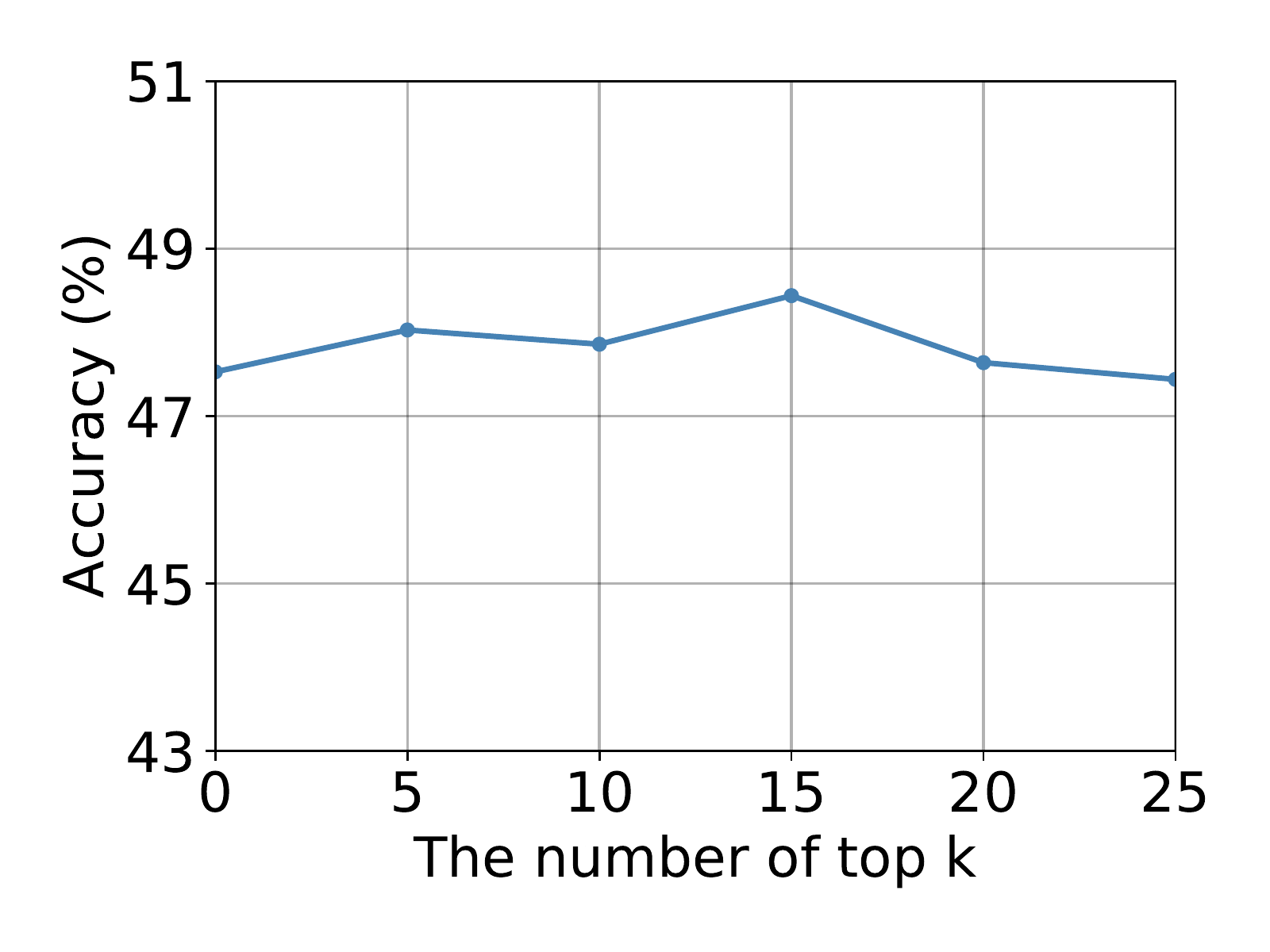}  
	}
	\subfigure[Squirrel] 
	{
		\centering      
		\includegraphics[width=0.45\columnwidth]{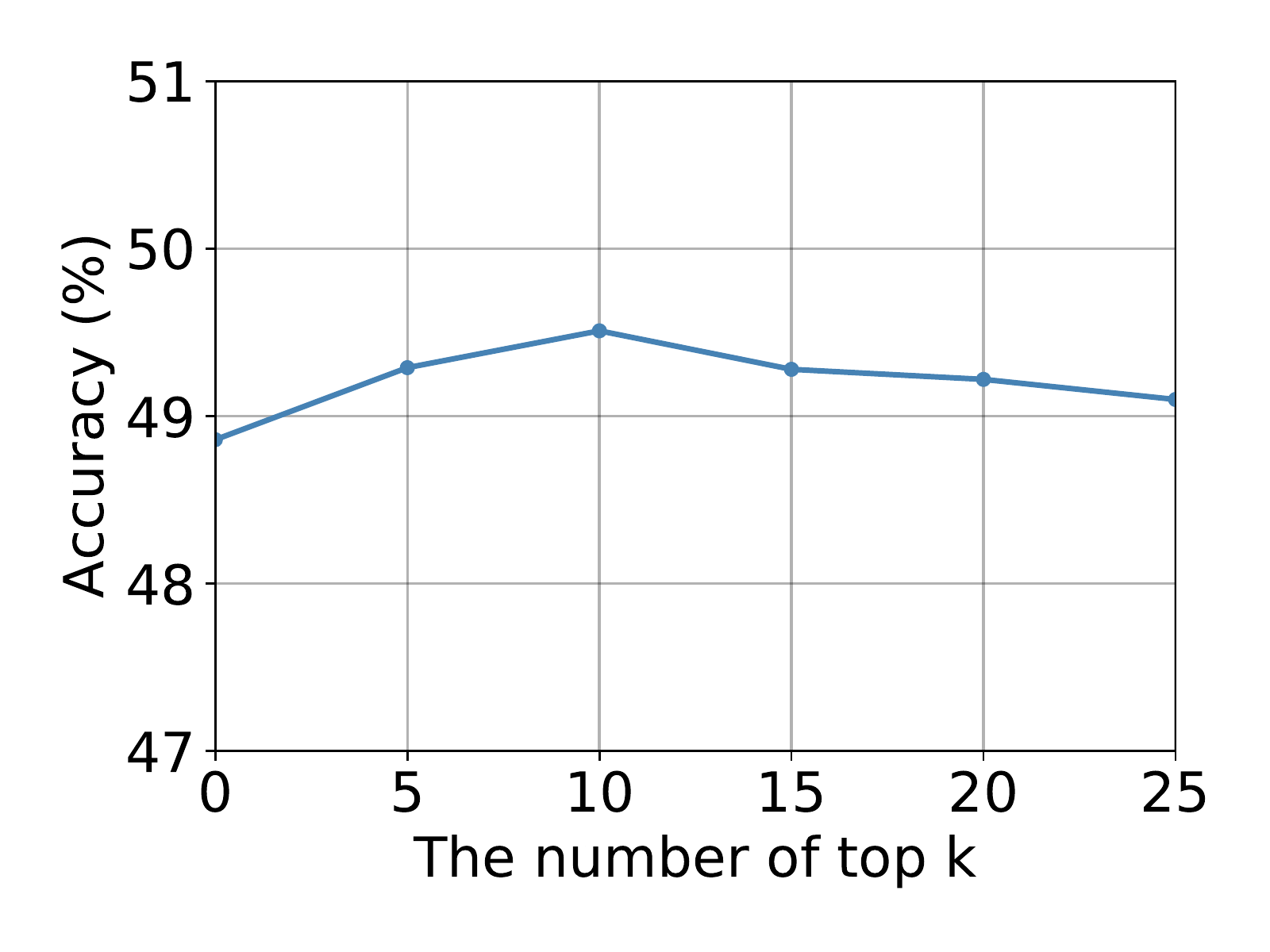}   
	}
	
	\caption{Parameter sensitivity analysis of top $k$.}\label{fig:k}
\end{figure}

\subsection{Ablation Study} 
We compare T2-GNN with its four variants on all datasets to intuitively understand the effectiveness of each component.
\begin{itemize}
\item w/o\_Teacher\_str: T2-GNN without structure teacher.
\item w/o\_Teacher\_fea: T2-GNN without feature teacher.
\item w/o\_Distill\_log: T2-GNN without logit distillation.
\item w/o\_Distill\_mid: T2-GNN without intermediate representation distillation.
\end{itemize}

\begin{figure}[h]  
	\centering       
	\includegraphics[width=0.95\linewidth]{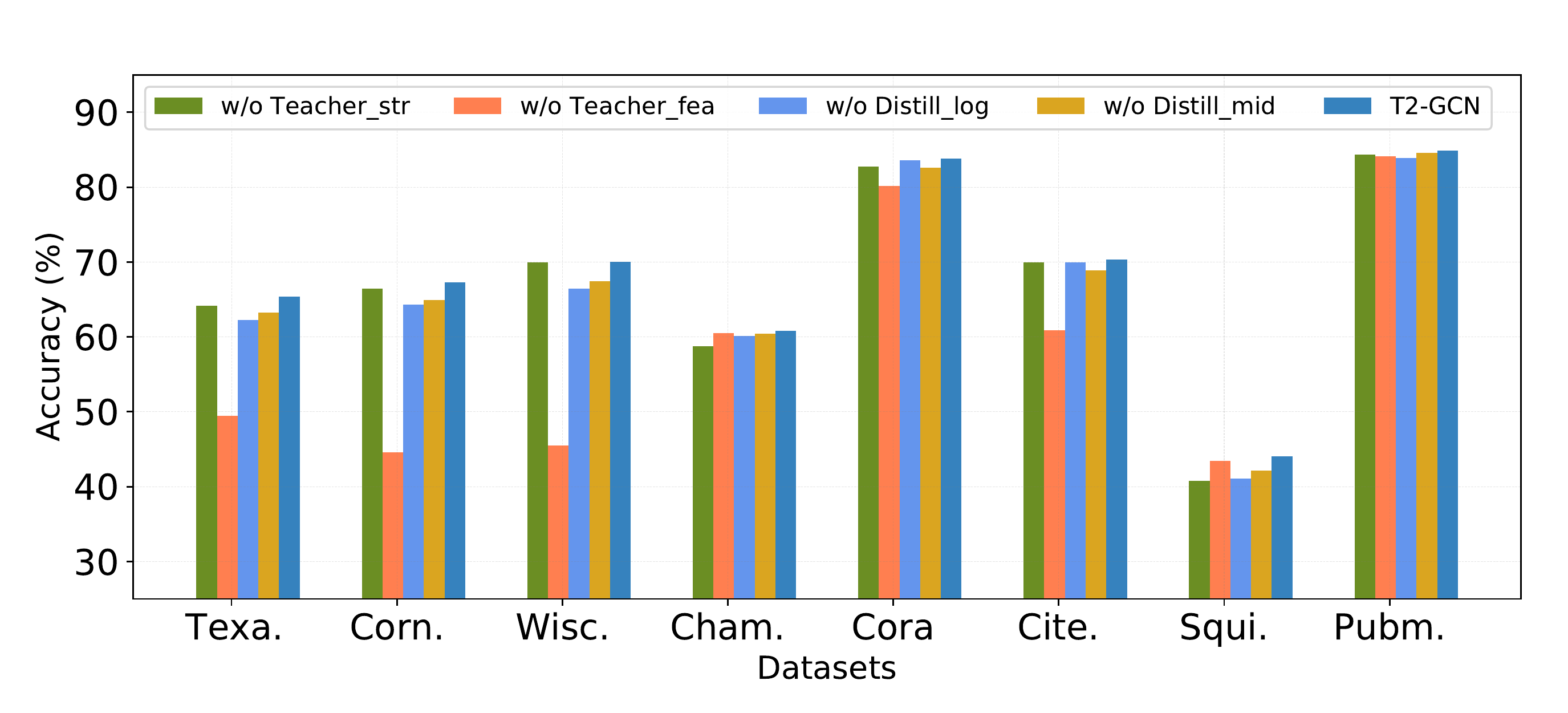} 
	\caption{Comparisons of the new T2-GNN with its four variants on node classification.}\label{fig:xr}  
\end{figure}

From the results in Fig.~\ref{fig:xr}, we can draw the following conclusions: (1) T2-GNN consistently outperforms all variants, indicating the effectiveness of using both feature and structure teachers as well as the dual distillation. (2) w/o\_Teacher\_fea outperforms  w/o\_Teacher\_str on the Chamelon and Squirrel datasets, but it is opposite on other datasets. This indicates that the degree of credibility of expert knowledge from features and structure in different datasets is different, so we need to weigh the weights of feature teacher and structure teacher to better provide targeted guidance (which is solved in our paper by balancing parameter $\lambda$). (3) Both w/o\_Distill\_log and w/o\_Distill\_mid achieve good performance but are lower than T2-GNN. This indicates that the dual distillation mechanism more effectively and comprehensively realizes the knowledge transfer from teacher model to student model.

\begin{figure}[t]  
	\centering    
	\subfigure[Wisconsin] 
	{
		\centering          
		\includegraphics[width=0.43\columnwidth]{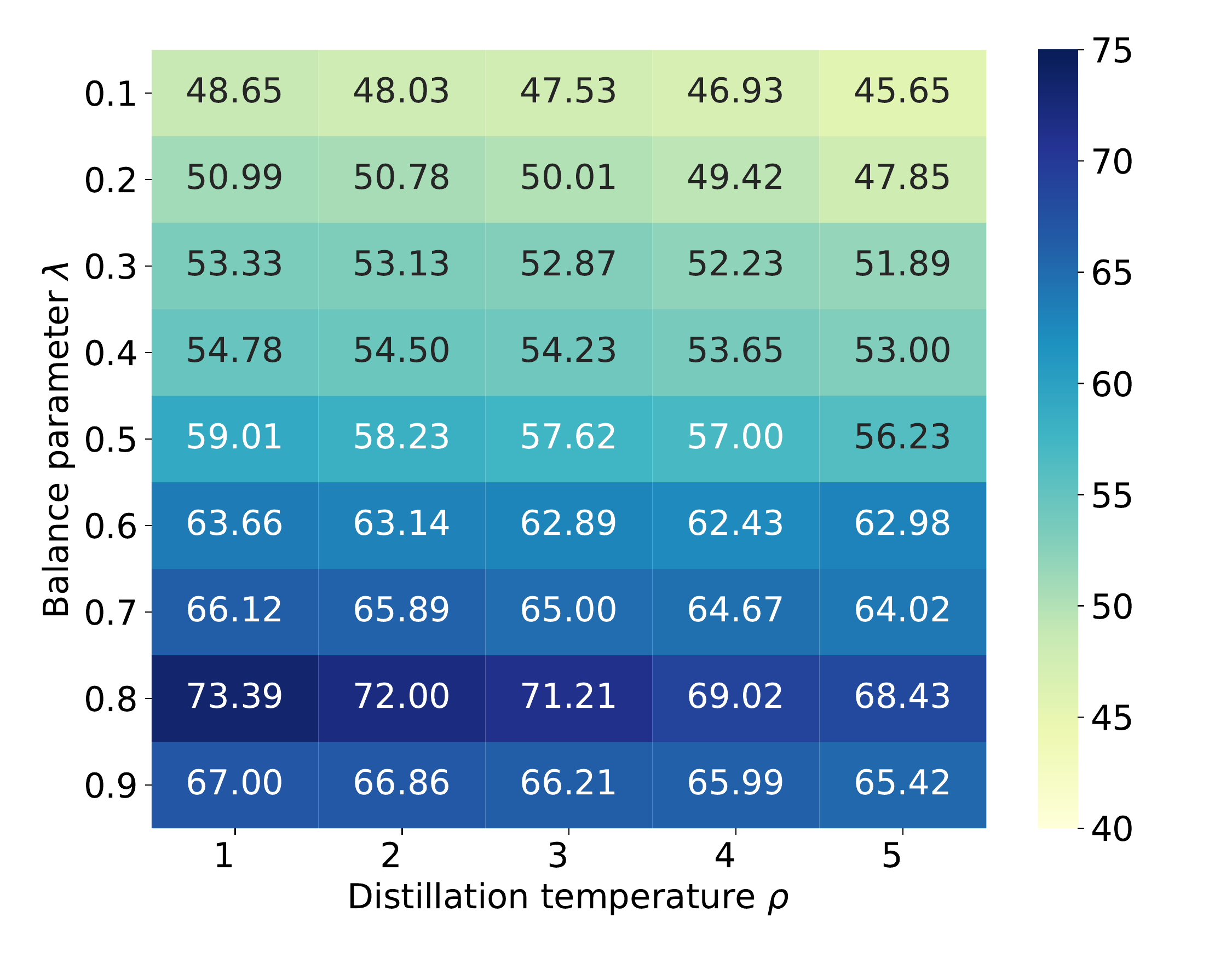}  
	}
	\subfigure[Squirrel] 
	{
		\centering      
		\includegraphics[width=0.43\columnwidth]{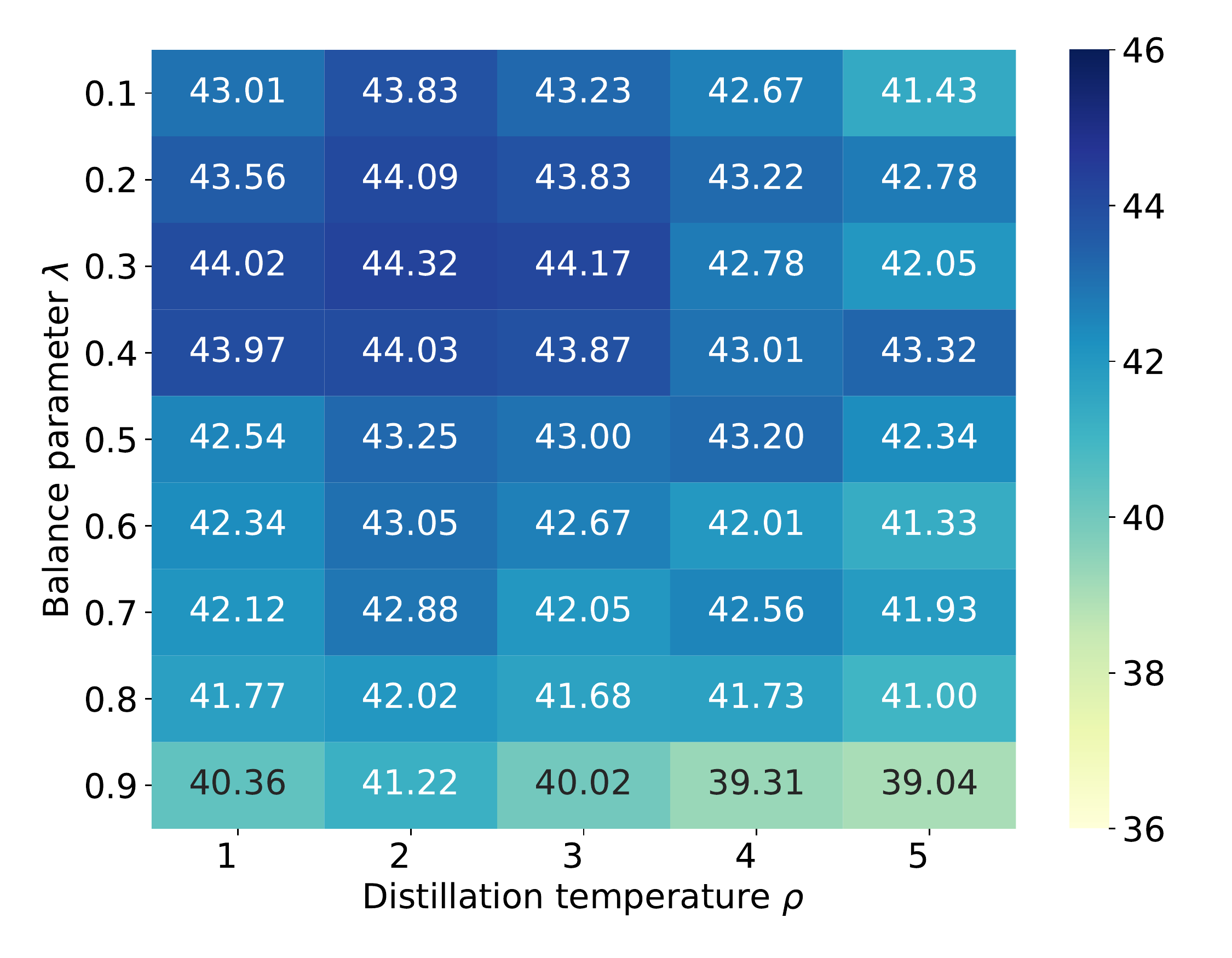}   
	}
	
	\caption{Parameter sensitivity analysis of $\lambda$ and $\rho$. }\label{fig:lamuda}
\end{figure}
\vspace{-10pt}
\subsection{Parameter Analysis}
Taking datasets Wisconsin and Squirrel of different scales as examples, we investigate the sensitivity of hyperparameters in T2-GNN.

\textbf{Analysis of top $k$}. The parameter $k$ in our framework refers to the number of PPR-based neighbor nodes per node in the structure enhancement. We vary it from 0 to 25. As shown in Fig.~\ref{fig:k},  the performance degraded when $k$ was large or small. This observation indicates that a small number of structurally enhancing relationships is not sufficient to obtain informative node embeddings, whereas too many may introduce noise and weaken the information propagation. 

\textbf{Analysis of balance parameter $\lambda$}.
In our framework, $\lambda$ is used to balance the weight of distillation loss from feature and structure teachers. Here we investigate the influence of $\lambda$ on node classification by varying it from 0.1 to 0.9 with a step size of $0.1$. As shown in Fig.~\ref{fig:lamuda}, T2-GNN achieves the best performance when $\lambda =0.8$ on Wisconsin and $\lambda =0.4$ on Squirrel. Its performance shows a trend of first rising and then descending and reaches its best at a certain value. This observation indicates that  node features and graph structure may have different degrees of trustworthiness when modeling incomplete graphs, because we cannot determine which missing information is more important to the node, as is the case in the real-world graph.

\textbf{Analysis of distillation temperature $\rho$}. In our framework, $\rho$ is a temperature parameter in logit-based distillation loss. The larger the $\rho$, the model training will pay more attention to the negative labels. We also investigate the influence of $\rho$ on node classification by varying it from 1 to 5. As Fig.~\ref{fig:lamuda} shows, $\rho$ is relatively insensitive and a small $\rho$ (e.g., 1 or 2) has a slight advantage. This observation indicates that due to the incompleteness of the graph itself, in order to avoid the influence of noise in the negative label, we need a relatively small distillation temperature.

\section{Conclusion}

In this paper, we propose a general GNN framework based on teacher-student distillation, which can effectively model incomplete graphs and avoid the interference between features and structure. In specific, we design feature-based and structure-based teacher models to capture effective information in features and structure, respectively. Then with the well-trained teacher model, targeted guidance is provided for the student model (base GNN, such as GCN) from the feature level and the structure level respectively. Finally, the dual distillation mode is adopted to achieve the comprehensive and effective transfer of knowledge from teachers to students. Extensive experiments demonstrate that the proposed new framework can effectively improve the performance of GNN on incomplete graphs, and further comparisons with baselines demonstrate its robustness.

\section{Acknowledgments}
This work was supported by the National Natural Science Foundation of China (62276187, 62272340, 62172056, 62176119).

\bibliography{aaai23}

\end{document}